\documentclass[runningheads]{llncs}

\usepackage{amsmath}
\usepackage{amsfonts}
\usepackage{amssymb}
\usepackage{subcaption}
\usepackage[ruled,vlined]{algorithm2e}
\usepackage{mathtools}
\usepackage{booktabs}
\usepackage{multirow}
\usepackage{makecell}
\usepackage{tabularx}
\usepackage{booktabs}
\usepackage{booktabs}
\usepackage{array}
\usepackage{bm}
\usepackage[table]{xcolor}
\newcolumntype{Y}{>{\centering\arraybackslash}X}
\definecolor{lightblue}{RGB}{200,225,255}

\usepackage{pifont}
\usepackage{colortbl}
\usepackage{xcolor}
\newcommand{\cmark}{\textcolor{green!60!black}{\ding{51}}}
\newcommand{\xmark}{\textcolor{red}{\ding{55}}}

\newcommand{\ms}[2]{#1\,{\tiny$\pm$\,#2}}
\newcommand{\msb}[2]{\textbf{#1}\,{\tiny$\bm{\pm}$\,\textbf{#2}}}

\usepackage[T1]{fontenc}
\usepackage{graphicx,verbatim}

\begin{document}
\title{Generative Brownian Bridge Diffusion In Motion Space For Enhanced Myocardial Strain Analysis}

\author{Rishov Paul\inst{1} \and
Frederick H. Epstein\inst{3} \and
Miaomiao Zhang\inst{1,2}}
\authorrunning{R. Paul et al.}
\institute{Department of Computer Science, University of Virginia, USA \and Department of Electrical and Computer Engineering, University of Virginia, USA \and Department of Biomedical Engineering, University of Virginia, USA}

\maketitle

\begin{abstract}
Myocardial strain analysis of cardiac magnetic resonance (CMR) images provides an important tool for evaluating cardiac function. However, current techniques require either human-adjusted post-processing with suboptimal regional accuracy, or specialized acquisitions with limited availability. In this paper, we propose to leverage the power of generative models to synthesize high-quality motion-derived strain values from routinely acquired CMR sequences. Specifically, we develop a novel Brownian bridge diffusion model in motion space to learn the probabilistic mapping between standard CMR motion estimated from widely adopted registration methods and highly accurate motion provided by advanced strain imaging techniques. To promote the fidelity of anatomical structure in the generation process, our model is conditioned on the corresponding CMR images. We validate our method on large-scale multi-center CMR datasets including subjects of paired standard cine CMR and advanced strain imaging acquisitions. Experimental results demonstrate that our framework significantly improves the accuracy of motion prediction and strain analysis from standard CMRs compared to existing learning-based approaches. Our research represents a new paradigm for potentially developing cost-effective, clinically deployable AI tools for cardiac function assessment with enhanced strain accuracy in busy clinical workflows. Our code is publicly available at \textit{https://github.com/Rishov-MIA/Brownian-Bridge-strain-analysis}.


\end{abstract}

\section{Introduction}
Quantification of myocardial strain from cine CMR image sequences plays an important role in assessing cardiac ventricular function and diagnosing heart diseases~\cite{amzulescu2019myocardial,seo2013computing}. Current approaches often rely on post-processing image registration~\cite{vercauteren2010symmetric,sotiras2013deformable} to track myocardial motion that provides a direct measure for strain values throughout the cardiac cycle. However, these methods are limited by heavy dependence on image texture and susceptibility to noise with compromised regional registration accuracy~\cite{sotiras2013deformable}. Advanced imaging techniques, such as displacement encoding with stimulated echoes (DENSE), have gained increasing attention and are emerging as a gold standard for myocardial strain analysis due to their ability to directly encode highly accurate tissue displacement/motion~\cite{zhong2010imaging,ghadimi2023improved}. Despite these advantages, there has not been widespread adoption of DENSE in routine clinical workflows due to limited availability of the methods and the need for additional acquisitions.

To bridge this gap, earlier studies have made substantial progress in developing deep learning–based image registration to improve the accuracy of myocardial motion and strain estimation from standard cine CMR images. These approaches incorporate spatial and temporal regularization to enforce motion continuity and stabilize the subsequent strain quantification~\cite{balakrishnan2018unsupervised,qiao2020temporally,morales2021deepstrain}. In parallel, architectural improvements, including multi-scale feature learning~\cite{ronneberger2015u} and transformer-based self-attention~\cite{vaswani2017attention}, have been introduced to better capture motion patterns across spatial or temporal resolutions. More recent efforts further integrate biomechanical priors and physics-based constraints to improve anatomical plausibility and physiological consistency of predicted motion fields~\cite{dalca2019unsupervised,qin2023generative}. Additionally, supervised paradigms have also emerged by leveraging advanced imaging modalities (i.e., DENSE) to provide high-fidelity motion supervision during training~\cite{wang2023strainnet,xing2024lamod}. While these methods have achieved state-of-the-art performance in motion and strain estimation, they remain limited in modeling the inherently multimodal nature of cardiac motion, particularly under pathological variability. This limitation comes from either deterministic modeling assumptions~\cite{hatamizadeh2022unetr,wang2023strainnet} or diffusion-based frameworks that primarily operate through iterative latent feature refinement rather than explicitly modeling the multimodal motion distribution~\cite{xing2024lamod}. Furthermore, many supervised approaches require spatiotemporal pre-alignment between paired acquisitions~\cite{hatamizadeh2022unetr,wang2023strainnet}, which can introduce artifacts and reduce model robustness.

Inspired by recent advances in Brownian bridge diffusion models~\cite{li2023bbdm}, we introduce a new generative diffusion framework in motion space to learn a stochastic mapping between paired motion representations. Specifically, our model is trained to map motion predicted from registration networks into high-quality advanced imaging modality; thereby serving as a refinement module of myocardial motion estimated from CMRs. In addition, we incorporate CMR image features as conditional guidance during the reverse diffusion process to ensure the generated motion trajectories remain consistent with underlying cardiac anatomy and temporal dynamics. Our main contributions are summarized as
\begin{enumerate}
 \renewcommand{\labelenumi}{(\roman{enumi})}
 \item Propose the first Brownian bridge diffusion framework operating directly in spatiotemporal displacement space to learn probabilistic mappings between readily obtainable registration-derived motion and difficult-to-acquire high-fidelity motion distributions from advanced imaging.
 \item Introduce structure-conditioned diffusion denoising that integrates CMR image features during the reverse diffusion process to enforce biomechanically and anatomically plausible spatiotemporal motion evolution.
 \item Demonstrate consistent and significant improvements over state-of-the-art methods in both motion reconstruction fidelity and subsequent myocardial strain quantification across a large multi-center dataset.
 \end{enumerate}

\section{Background: Registration-based Motion Estimation} 
\label{sec:background}

In this section, we will review a general mathematical framework for motion estimation based on deformable image registration~\cite{wu2024tlrn,chandrashekara2003construction,pan2021efficient}. The goal is to estimate displacement fields that describe myocardial motion over time by aligning a temporal image sequence to a reference frame.

Let $\Omega \subset \mathbb{R}^d$ denote a $d$-dimensional image domain. Given a reference image $I_0:\Omega \rightarrow \mathbb{R}$ and a sequence of moving images $\{I_t\}_{t=1}^{T}, t \in \{1, \ldots, T\}$, the objective is to estimate a sequence of spatial transformations, $\{\phi_t\}_{t=1}^{T} : \Omega \rightarrow \Omega$, that map spatial locations from the reference frame to the target frame at time $t$. The corresponding displacement field is defined as $u_t = \phi_t - e$, where $e$ denotes an identity transformation. A classical deformable registration~\cite{sotiras2013deformable,beg2005computing,arsigny2006log} formulates motion estimation as an energy minimization problem over a sequence of transformations $\{\phi_t\}$, parameterized by velocity fields $v_t$~\cite{arsigny2006log,arnold1966,beg2005computing}, as
\begin{equation}
\text{min} \, \mathcal{L}(\phi_t) = \sum_{t=1}^{T}
\mathrm{Dist}\!\left(I_0 \circ \phi_t^{-1},\, I_t \right)
+ \lambda \, \mathrm{Reg}(v_t),
\label{eq:temporal_registration_energy}
\end{equation}
where $\circ$ denotes an interpolation operator, $\mathrm{Dist}(\cdot,\cdot)$ measures image dissimilarity, and $\mathrm{Reg}(\cdot)$ is a regularization term that enforces the spatiotemporal smoothness of the transformation fields. The parameter $\lambda>0$ controls the trade-off between image matching fidelity and deformation regularity. 

Various choices of spatial regularizations, such as stationary velocity field (SVF)~\cite{arsigny2006log} and large deformation diffeomorphic metric mapping (LDDMM)~\cite{arnold1966,beg2005computing}, enforce diffeomorphic constraints (a.k.a. one-to-one smooth and inverse smooth mapping) to ensure anatomically plausible transformations. In particular, under the SVF parameterization, the transformation is obtained from the velocity field through the exponential map, i.e., $\phi_t = \exp(v_t)$, and its inverse is directly given by $\phi_t^{-1} = \exp(-v_t)$. In this paper, we adopt commonly used sum-of-squared intensity differences (SSD) as the similarity metric~\cite{wu2024tlrn,rueckert2006diffeomorphic}, and SVF for the regularity. 

Once displacement fields are estimated, myocardial strain can be computed from the local deformation gradient tensor~\cite{xing2024lamod,barnafi2024reconstructing}, i.e., $F_t = \nabla \phi_t = e + \nabla u_t$, where $\nabla u_t$ is the Jacobian of the displacement field. For finite deformation analysis, myocardial strain is commonly quantified using the Green-Lagrange strain tensor~\cite{smiseth2016myocardial,marwick2019myocardial}, i.e.,
\begin{equation}
\label{eq:strain}
S_t = \frac{1}{2}\big(F_t^{T} F_t - e \big).
\end{equation}

\section{Our Method}
\begin{figure}[!t]
    \centering
    \includegraphics[width=\textwidth]{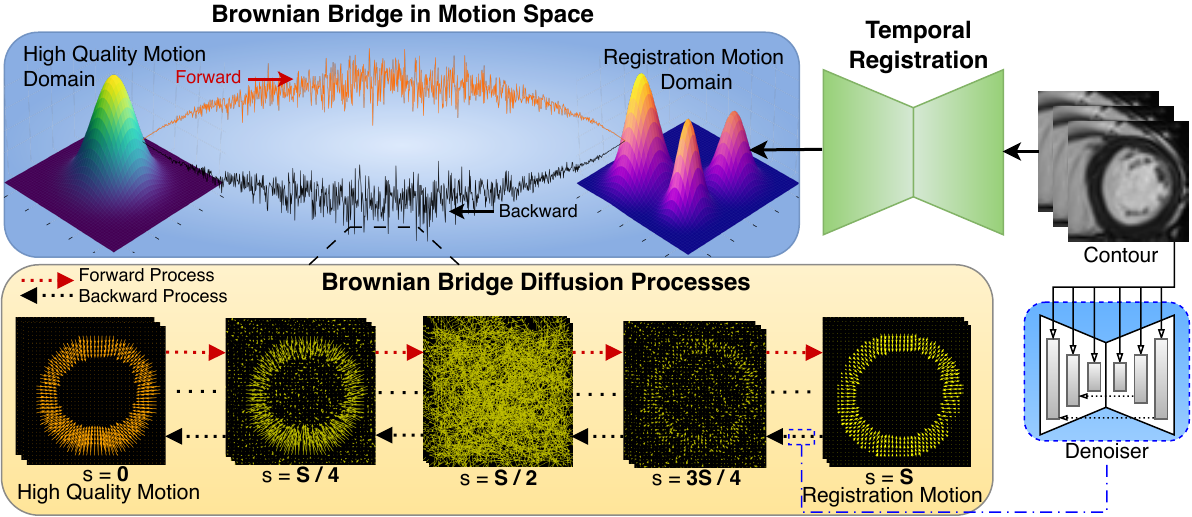}
    \caption{An overview of our Brownian bridge diffusion model in motion space.}
    \label{fig:architecture}
\end{figure}

In this section, we develop a conditional Brownian bridge diffusion model in a vectorized spatiotemporal displacement space to explicitly learn the probabilistic mapping between two distinct motion domains. Once trained, the proposed model takes motion fields estimated from standard cine CMR sequences and translates them through the learned diffusion bridge to generate high-fidelity motion estimates without specialized acquisition. Our framework includes two key components: (i) a temporal registration network that predicts spatiotemporal motion fields from input CMR videos, and (ii) a conditional Brownian bridge video diffusion model that synthesizes high-quality motion trajectories given previously learned registration-derived motion. An overall architecture of our proposed method is shown in Fig.~\ref{fig:architecture}.

\paragraph*{\bf Motion learning from spatiotemporal registration network.}
Given a training dataset of $N$ cardiac image sequences, $\{\mathcal{I}^i\}_{i=1}^N, i \in \{1, \cdots, N\}$, where each sequence $\mathcal{I}^i$ contains $T+1$ temporal frames, i.e., $\mathcal{I}^i = \{ \mathcal{I}^{i}_{t} \}_{t=0}^T$. Our objective is to predict a sequence of velocity fields, $\{v^i_t\}$, from which the deformation fields $\{\phi^i_t\}$ associated with the reference image and following images $\{\mathcal{I}^i_0, \mathcal{I}^i_t\}_{t=1}^{T}$ are obtained via the SVF regularization~\cite{arsigny2006log}. Analogous to~\cite{wu2024tlrn}, we employ a spatiotemporal registration network parameterized by $\theta$, and formulate the corresponding loss function as
\begin{equation}
\mathcal{L}(\theta)
=
\sum_{i=1}^{N}
\Bigg[
\sum_{t=1}^{T}
\left\|
\mathcal{I}_0^i \circ \phi_t^i(\theta)-\mathcal{I}_t^i
\right\|_2^2
+ \sum_{t=1}^{T}\lambda\left\|\nabla v^i_t(\theta)\right\|^2
+ \sum_{t=2}^{T}\beta\left\|\Delta v^i_t(\theta)\right\|^2
\Bigg],
\end{equation}
where $\Delta v^i_t(\theta) \triangleq v^i_t(\theta) - v^i_{t-1}(\theta)$, representing a temporal regularization, and $\beta$ is a controlling weighting parameter. Given the predicted transformation fields $\{\phi_t^i\}$, the displacement fields can be computed as $u_t^i = \phi_t^i - e$. Note that, in this paper, we adopt a U-Net backbone with a latent residual network to explicitly model temporal correlations in the registration framework~\cite{wu2024tlrn}. Our proposed methodology is architecture-agnostic, and alternative backbone designs~\cite{chen2022transmorph,balakrishnan2019voxelmorph} can be readily incorporated without modification to the overall formulation.

\paragraph*{\bf Conditional Brownian Bridge Diffusion in Motion Space.}

Let $(\mathbf{x}, \mathbf{u} ) \in \mathbb{R}^{d \times T \times H \times W}$ denote paired high quality motion acquired by advanced imaging and registration-derived motion fields in the same spatial coordinate system using the same reference image. Here, $H$ and $W$ denote the height and width of each image. We model motion refinement as a stochastic Brownian bridge defined over the motion space, with boundary conditions
$\mathbf{x}_0 = \mathbf{x}$ and $\mathbf{x}_S = \mathbf{u}$, where $S$ represents the ending point. 

Following a similar principle of~\cite{li2023bbdm}, the {\em forward diffusion process} evolves intermediate motion states by progressively transporting $\mathbf{x}$ toward $\mathbf{u}$ via a variance-preserving linear interpolation. Note that since the paired motions are defined with respect to the same reference image, we assume the interpolation is performed without introducing coordinate misalignment or reparameterization errors. Let $m_s \in [0,1]$ denote a monotonically increasing schedule for $s \in \{0,\dots,S\}$. The forward process in the motion space can be formulated as 
\begin{equation}
q_{\mathrm{BB}}(\mathbf{x}_s \mid \mathbf{x}, \mathbf{u})
=
\mathcal{N}\!\left(
(1 - m_s)\mathbf{x} + m_s \mathbf{u},\,
\delta_s \mathbf{I}
\right);
\quad
m_s = s/S.
\end{equation}
where $\delta_s = 2\tau(m_s - m_s^2)$ with $\tau$ being the variance scaling factor~\cite{li2023bbdm}. Sampling from this process gives
$\mathbf{x}_s=(1 - m_s)\mathbf{x}+m_s \mathbf{u}+\sqrt{\delta_s}\,\boldsymbol{\epsilon}$, where
$\boldsymbol{\epsilon} \sim \mathcal{N}(\mathbf{0},\mathbf{I}).$

The {\em reverse process} is initialized by enforcing the ending boundary condition $\mathbf{x}_S = \mathbf{u}$. Starting from this fixed endpoint, the model progressively refines the registration-motion trajectory toward the high-quality motion domain. To improve data fidelity and enforce anatomical consistency, we condition the reverse dynamics on image features, denoted as $\mathbf{c}$, derived from the input data (e.g., myocardial contours or latent CMR image representations). Formally, our reverse transition can be parameterized as a conditional Gaussian 
\begin{equation}
p_\theta(\mathbf{x}_{s-1} \mid \mathbf{x}_s, \mathbf{u}, \mathbf{c}) 
= 
\mathcal{N} (
\boldsymbol{\mu}_\theta(\mathbf{x}_s, s, \mathbf{c}),\, 
\tilde{\delta}_s \mathbf{I}
),
\end{equation}
where $\mu_\theta$ is predicted by a neural network (e.g., with a commonly used 3D UNet backbone) and variance $\tilde{\delta}_s$ follows a fixed variance schedule~\cite{li2023bbdm}. 

Similar to~\cite{li2023bbdm}, instead of predicting the noise directly, we parameterize the denoising objective using a bridge gradient formulation defined as $\boldsymbol{b}_s = m_s(\mathbf{u} - \mathbf{x}) + \sqrt{\delta_s}\boldsymbol{\epsilon}.$ At each timestep, the denoising network predicts the Brownian bridge update along the motion trajectory, i.e., $\hat{\boldsymbol{b}}_s = f_\theta(x_s, s, c)$, from which the  displacement field is recovered as  $\hat{\mathbf{x}}_0 = \mathbf{x}_s - \hat{\mathbf{b}}_s$. To explicitly enforce physically plausible and smooth motions, we incorporate a spatial regularization into the loss function to constrain non-smooth or anatomically implausible updates. The network parameters $\theta$ are then learned by minimizing loss function as
\begin{equation}
\label{eq:cost_function}
\min_{\theta}\ \mathcal{L}(\theta)
=
\mathbb{E}_{\mathbf{x}_0,u,\boldsymbol{\epsilon},\mathbf{c},s}
\Big[
\left\|
\mathbf{b}_s - \hat{\mathbf{b}}_s
\right\|
+\lambda_{\mathrm{reg}} \sum\nolimits_{t=1}^{T}\|\nabla \hat{\mathbf{x}}_0^{\,t}\|_2^2
\Big],
\end{equation}
where $\lambda_{\mathrm{reg}}$ is a regularization parameter. In this paper, we also incorporate segmented CMR myocardium contours as conditioning information into each motion state $\mathbf{x}_s$ through a multi-head cross-attention mechanism~\cite{vaswani2017attention}.

During inference, we initialize the reverse process directly with the registration motion $\mathbf{x}_S = \mathbf{u}$ 
and iteratively sample 
$\mathbf{x}_{s'} = (1 - m_{s'})\hat{\mathbf{x}}_0 + m_{s'}\mathbf{u} 
+ \sqrt{(\delta_{s'} - \tilde{\delta}_s)/\delta_s}\,
(\mathbf{x}_s - (1 - m_s)\hat{\mathbf{x}}_0 - m_s \mathbf{u}) 
+ \sqrt{\tilde{\delta}_s}\,\boldsymbol{\epsilon}$, 
where $s' < s$ is the next retained timestep in a subsampled non-Markovian schedule and $\tilde{\delta}_s = \delta_{s'}/\delta_s(\delta_s - \delta_{s'}(1-m_s)^2/(1-m_{s'})^2)\,$. At the final timestep $s=0$, the model directly returns the reconstructed motion $\hat{\mathbf{x}}_0$ without noise injection.

\section{Experimental Evaluation}
We compare our model against state-of-the-art methods for myocardial strain estimation, covering both deterministic (TransUNet~\cite{chen2021transunet}, and StrainNet~\cite{wang2023strainnet}) and generative approaches (Flow-matching~\cite{tong2023improving}, vid2vid~\cite{wang2018vid2vid}, ControlNet~\cite{zhang2023adding}, and LaMoD~\cite{xing2024lamod}). For a fair comparison, all baselines were retrained from scratch, and every model was trained with Adam~\cite{Kingma2014Adam} on a single NVIDIA A100 80GB GPU. The hyperparameters of our model were selected via grid search, resulting in a batch size of $16$, learning rate and weight decay of $1\times 10^{-4}$, spatial regularization weight $\lambda_{\mathrm{reg}} = 0.01$, and $\tau = 1$ for all experiments.

\noindent \textbf{Dataset.} We include two CMR modalities: routinely acquired left-ventricular (LV) short-axis cine CMR sequences and advanced DENSE-CMR acquisitions, along with manually annotated LV masks~\cite{xing2024lamod,wang2023strainnet}. The DENSE-CMR provides both magnitude images and high-quality displacement/motion measurements that serve as a reference standard for circumferential strain evaluation. The DENSE dataset was collected from eight clinical centers~\cite{wang2023strainnet} 
and consists of $1{,}200$ sequences from $284$ subjects ($124$ healthy volunteers and $160$ patients with heart disease). Sequences are divided subject-wise into training, validation, and testing sets at an $80{:}10{:}10$ ratio to prevent data leakage across splits. Each of the $116$ test sequences is paired with a cine acquisition from the same subject at approximately the same slice location (within $\pm4$\,mm) which lets the DENSE-derived displacements serve as ground-truth motion for the cine sequences. All CMR videos are cropped to a $48^2\times 20$ LV region of interest, with average in-plane spatial resolution being 2.41\,mm/pixel for DENSE and 1.68\,mm/pixel for cine. Our model is trained exclusively on DENSE-CMR and evaluated on both DENSE and the paired cine CMR sequences.

\noindent \textbf{Evaluation.} We first evaluate the accuracy of predicted displacement vs. DENSE ground truth using a normalized pixel-wise end-point error (nEPE)~\cite{wang2023strainnet,xing2024lamod}, defined as the EPE normalized by the ground-truth displacement magnitude. Our model takes DENSE magnitude images as input, and its predicted displacement fields are compared with ground-truth fields computed from the corresponding DENSE phase images. We report nEPE on DENSE-CMR, where magnitude images and displacement fields share the same spatial coordinate system.

We then compute myocardial circumferential strain from the motions predicted from cine CMR images using Eq.~\eqref{eq:strain}, and compare our method against all baselines using mean absolute error (MAE) under four evaluation settings: whole-slice and segmental strain, each across all frames and at end-systole~(ES), defined as the frame of minimum strain value. Strain MAE is reported in percentage points (\%), representing absolute error rather than relative percent difference. Note that myocardial strain values typically range up to approximately 25\% rather than 100\%~\cite{wang2023strainnet,morales2021deepstrain}.
Following the standard clinical convention~\cite{amzulescu2019myocardial}, the myocardium is partitioned into six segments for basal and mid-ventricular slices and four segments for apical slices, beginning at the right ventricular insertion point and proceeding counterclockwise. Whole-slice strain is obtained by averaging the segmental strain values prior to error computation. All predicted strain measurements are compared against reference strain values derived from the paired high-quality DENSE. Due to space limit, we report significance on segmental ES MAE using two-sided
Wilcoxon signed-rank tests~\cite{wilcoxon1992individual} against each baseline, with Holm correction~\cite{holm1979simple} for multiple comparisons ($p<0.05$).

\begin{figure}[!t]
    \centering
    \includegraphics[width=\textwidth]{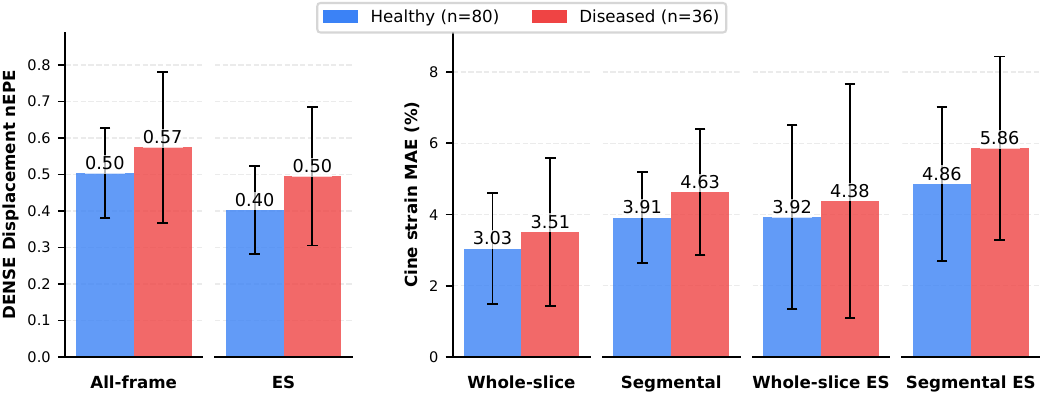}
    \caption{Left to right: errors of displacements and strain on stratified cohorts.}
    \label{fig:stratified_results}
\end{figure}

\noindent{\textbf{Experiment Results.}}
Tab.~\ref{tab:disp_strain_error_metrics_dense_cine} reports both DENSE displacement errors and cine-derived strain errors on paired Cine-DENSE test data. The results demonstrate that our method achieves top-tier motion tracking fidelity on par with the strongest baseline (LaMoD~\cite{xing2024lamod}) in displacement nEPE. Similarly, our method consistently offers the best performance for cine-derived strain estimation across all evaluation settings, with particularly notable improvements in ES strain analysis. Since myocardial strain spans a narrow range of roughly 25\%, a strain MAE of 1\% already corresponds to about 4-5\% of the clinically relevant range. In Tab.~\ref{tab:disp_strain_error_metrics_dense_cine}, the most substantial gains are observed in ES strain estimation, which underscore our method's ability to accurately capture peak myocardial deformation. The Holm-corrected Wilcoxon signed-rank tests confirm that the improvement in segmental ES strain is significant against every baseline ($p<0.05$). Furthermore, the lower standard deviations of errors suggest the enhanced robustness and stability of our approach over all baselines. Fig.~\ref{fig:stratified_results} reports the performance of our method across stratified cohort (healthy vs. diseases). It shows that errors (both on displacements and strain values) are generally higher on diseased subjects where more heterogeneous motion exhibits.

\begin{table}[htbp]
\begingroup
\centering
\caption{Comparison of displacement/strain error on paired Cine--DENSE data.}
\label{tab:disp_strain_error_metrics_dense_cine}
\setlength{\tabcolsep}{1pt}
\resizebox{\textwidth}{!}{%
\begin{tabular}{@{}l*{7}{c}@{}}
\toprule
\multirow{2}{*}{\textbf{Method}} &
\multicolumn{2}{c}{\makecell{\textbf{DENSE Displacement}\\\textbf{nEPE} $\downarrow$}} &
\multicolumn{4}{c}{\makecell{\textbf{Cine Strain}\\\textbf{MAE} (\%) $\downarrow$}} &
\multirow{2}{*}{\makecell{\textbf{Segmental}\\\textbf{ES}\\$p$\textbf{-value}}} \\
\cmidrule(lr){2-3}\cmidrule(lr){4-7}
 & \textbf{All-frame} & \textbf{ES}
 & \makecell{\textbf{Whole-}\\\textbf{slice}}
 & \textbf{Segmental}
 & \makecell{\textbf{Whole-}\\\textbf{slice ES}}
 & \makecell{\textbf{Segmental}\\\textbf{ES}} & \\
\midrule
TransUNet     & \ms{0.60}{0.14} & \ms{0.52}{0.16} & \ms{5.37}{2.24} & \ms{5.88}{1.98} & \ms{7.84}{3.93}  & \ms{8.46}{3.42}  & $<$0.001 \\
StrainNet     & \ms{0.56}{0.14} & \ms{0.49}{0.13} & \ms{3.87}{1.76} & \ms{4.57}{1.40} & \ms{5.55}{3.05}  & \ms{6.37}{2.39}  & $<$0.001 \\
\midrule
vid2vid       & \ms{0.65}{0.20} & \ms{0.54}{0.21} & \ms{7.76}{3.12} & \ms{8.21}{2.82} & \ms{11.52}{4.33} & \ms{11.82}{3.83} & $<$0.001 \\
Flow-matching & \ms{0.53}{0.21} & \ms{0.44}{0.18} & \ms{3.96}{2.43} & \ms{5.22}{2.18} & \ms{5.26}{4.65}  & \ms{6.79}{4.03}  & 0.013 \\
ControlNet    & \ms{0.54}{0.19} & \ms{0.46}{0.19} & \ms{4.42}{2.29} & \ms{4.87}{2.06} & \ms{5.87}{3.77}  & \ms{6.53}{3.35}  & $<$0.001 \\
LaMoD         & \ms{0.53}{0.19} & \ms{0.44}{0.16} & \ms{3.69}{1.71} & \ms{4.22}{1.45} & \ms{5.03}{3.16}  & \ms{5.80}{2.68}  & 0.013 \\
\rowcolor{lightblue}
Ours & \msb{0.52}{0.16} & \msb{0.43}{0.15} & \msb{3.18}{1.75} & \msb{4.14}{1.49} & \msb{4.06}{2.83} & \msb{5.17}{2.35} & --- \\
\bottomrule
\end{tabular}}
\endgroup
\end{table}

\noindent\textbf{Ablation Study.} Tab.~\ref{tab:ablation} reports the ablation study on evaluating the individual contribution of Brownian bridge, as well as CMR feature conditioning in our proposed framework. Results show that incorporating the Brownian bridge module substantially improves performance, yielding significant reductions in both displacement and strain errors compared with the registration-only model. Further integrating anatomical contour guidance leads to additional performance gains and consistently achieves the best results across all evaluation metrics. 
\begin{table}[!b]
\scriptsize
\centering
\caption{Ablation study to assess each proposed module in our proposed method.}
\resizebox{\linewidth}{!}{
\begin{tabular}{cccccccc}
\toprule
\multicolumn{3}{c}{\textbf{Components}} &
\textbf{DENSE Displacement} &
\multicolumn{4}{c}{\textbf{Cine Strain MAE} (\%) $\downarrow$} \\
\cmidrule(lr){1-3}\cmidrule(lr){4-4}\cmidrule(lr){5-8}
\makecell{\textbf{Registra-}\\\textbf{tion}} &
\makecell{\textbf{Brownian}\\\textbf{Bridge}} &
\makecell{\textbf{Contour}\\\textbf{Guidance}} &
\textbf{nEPE} $\downarrow$ &
\makecell{\textbf{Whole-}\\\textbf{slice}} &
\textbf{Segmental} &
\makecell{\textbf{Whole-}\\\textbf{slice ES}} &
\makecell{\textbf{Segmental}\\\textbf{ES}} \\
\midrule
\cmark & \xmark & \xmark
& 0.73 {\tiny$\pm$ 0.18}
& 5.35 {\tiny$\pm$ 2.10}
& 6.01 {\tiny$\pm$ 1.87}
& 6.05 {\tiny$\pm$ 3.36}
& 6.89 {\tiny$\pm$ 2.86} \\
\cmark & \cmark & \xmark
& 0.62 {\tiny$\pm$ 0.18}
& 3.78 {\tiny$\pm$ 2.03}
& 4.78 {\tiny$\pm$ 2.01}
& 5.01 {\tiny$\pm$ 3.58}
& 5.94 {\tiny$\pm$ 3.15} \\
\rowcolor{lightblue}
\cmark & \cmark & \cmark
& \textbf{0.52 {\tiny$\pm$ 0.16}}
& \textbf{3.18 {\tiny$\pm$ 1.75}}
& \textbf{4.14 {\tiny$\pm$ 1.49}}
& \textbf{4.06 {\tiny$\pm$ 2.83}}
& \textbf{5.17 {\tiny$\pm$ 2.35}} \\
\bottomrule
\end{tabular}
}
\label{tab:ablation}
\end{table}

\begin{figure}[htbp]
    \centering
    \includegraphics[width=0.90\textwidth]{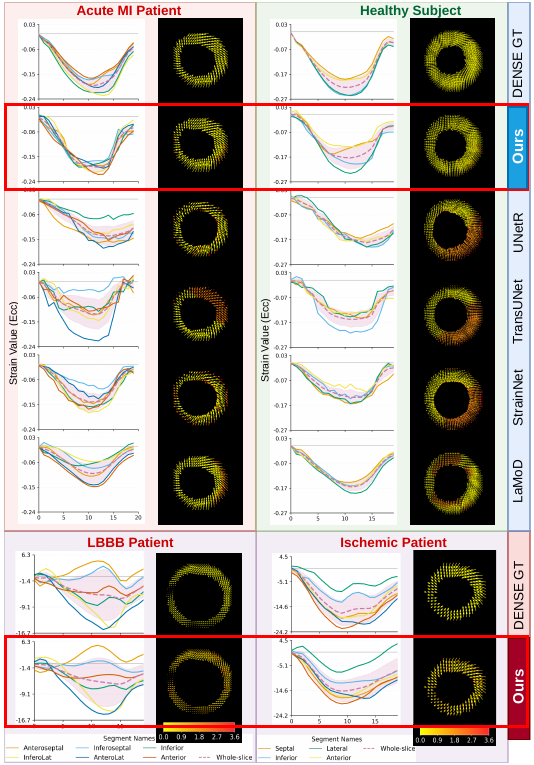}
    \caption{Top: a comparison of strain curves and ES displacement fields from DENSE ground truth vs. those predicted by all methods. Bottom: additional examples from LBBB and ischemic patients, comparing DENSE vs. predictions from our model.}
    \label{fig:examples}
\end{figure}

\noindent{\textbf{Qualitative Analysis.}} Fig.~\ref{fig:examples} visualizes four representative cases spanning varying pathologies. For an acute myocardial infarction (MI) patient and a healthy subject (top), we compare the DENSE ground truth (GT), our method, and the four best-performing baseline models reported from Tab.~\ref{tab:disp_strain_error_metrics_dense_cine}. The results are demonstrated using segmental strain curves and ES displacement fields overlaid with error maps. Our method achieves the highest agreement with the DENSE-strain across all segments throughout the cardiac cycle. Similarly, the ES displacement shows the highest similarity to DENSE (colored in yellow). In contrast, all baseline methods present noticeably larger deviations (colored in red). Due to space limitations, the additional LBBB and ischemic cases (bottom) show only our method against the DENSE GT, where the baselines follow the same trends reported above. Our model preserves the inter-segmental dyssynchrony characteristic of LBBB and the reduced deformation of the affected inferior segment in the ischemic case.

\noindent{\textbf{Discussion.}} While our method demonstrates improved agreement with DENSE-derived myocardial strain measurements, several limitations also point to promising directions for future work. First, our validation focuses on agreement with DENSE-derived displacement and strain, and does not yet establish whether the synthesized motion captures clinically discriminative patterns. Evaluating the generated motion on downstream tasks, such as disease classification, will be an important next step. Second, our model errors relatively higher in the diseased cohort indicate underrepresented heterogeneous motion in the training data. Improving training balance and representation of these challenging cases may further improve performance. Third, our method may inherit errors from registration networks when used as initial conditions for the Brownian bridge learning. This represents another promising direction for future work.

\section{Conclusion}
In this paper, we introduce the first Brownian bridge framework in the cardiac motion space to synthesize advanced yet clinically less accessible DENSE-quality motion from spatiotemporal registration-predicted displacement trajectories. In order to enhance the fidelity and anatomical consistency of the synthesized motion, we incorporate complementary CMR image features as an effective conditioning during the reverse process. Experimental results on a large multi-site cohort demonstrate that our approach consistently outperforms state-of-the-art myocardial strain analysis methods without requiring specialized acquisitions at inference. By bridging the gap between routine cine imaging and advanced myocardial motion acquisition, our framework provides a promising approach for improving myocardial motion and strain estimation from standard cine CMR, with the potential of future translation into clinical workflows.

\paragraph*{\bf Disclosure of Interests.} The authors have no conflicts of interest to disclose.

\paragraph*{\bf Acknowledgment.} This work was supported by NIH 1R21EB032597 and funding from the UVA Wallace H. Coulter Center for Translational Research.

\bibliographystyle{splncs04}
\bibliography{references}

\end{document}